\documentclass[runningheads]{llncs}
\usepackage{graphicx}

\usepackage{tikz}
\usepackage{comment}
\usepackage{amsmath,amssymb} 
\usepackage{color}

\usepackage[accsupp]{axessibility}  

\usepackage{floatrow}
\usepackage[font=small,labelfont=bf]{caption}
\usepackage{multicol}
\newfloatcommand{capbtabbox}{table}[][\FBwidth]

\begin{document}
\pagestyle{headings}
\mainmatter
\def\ECCVSubNumber{10}  

\title{Generative Poisoning Using Random Discriminators} 

%
\author{Dirren van Vlijmen\inst{1} \and
Alex Kolmus\inst{1}$^*$ \and
Zhuoran Liu\inst{1}$^*$ \and 
Zhengyu Zhao\inst{2} \and
Martha Larson\inst{1} }
\authorrunning{D. van Vlijmen et al.}
%
\institute{Radboud University, Nijmegen, The Netherlands \\ \email{\{dirren.vanvlijmen,alex.kolmus,zhuoran.liu,martha.larson\}@ru.nl}\and
CISPA Helmholtz Centre for Information Security, Saarbrücken, Germany\\ \email{zhengyu.zhao@cispa.de}}
\def\thefootnote{*}\footnotetext{Equal contribution}\def\thefootnote{\arabic{footnote}}
\maketitle
\begin{abstract}
We introduce ShortcutGen, a new data poisoning attack that generates sample-dependent, error-minimizing perturbations by learning a generator.
The key novelty of ShortcutGen is the use of a randomly-initialized discriminator, which provides spurious shortcuts needed for generating poisons.
Different from recent, iterative methods, our ShortcutGen can generate perturbations with only one forward pass in a label-free manner, and compared to the only existing generative method, DeepConfuse, our ShortcutGen is faster and simpler to train while remaining competitive.
We also demonstrate that integrating a simple augmentation strategy can further boost the robustness of ShortcutGen against early stopping, and combining augmentation and non-augmentation leads to new state-of-the-art results in terms of final validation accuracy, especially in the challenging, transfer scenario.
Lastly, we speculate, through uncovering its working mechanism, that learning a more general representation space could allow ShortcutGen to work for unseen data.
\end{abstract}
\section{Introduction}\label{Sec: Introduction}


The field of protective poisoning attacks has been gathering attention \cite{Fowl2021,Huang2021,Chen2020,Feng2019,Liu2021} with the rise of the non-consensual use of social images to train deep learning models \cite{Birhane2021,Amazon_NYtimes,China_NYtimes,ClearView_NYtimes}. 
Poisoning is conventionally achieved by injecting additional poisoning images \cite{Chen2020,Munoz2019} into the training set, and more recently, by slightly perturbing images already in the training set \cite{Feng2019,Fowl2021,Huang2021,Liu2021}, making it less noticeable~\cite{SHNSSDG18,SKPL20}.
Most existing studies on perturbation-based poisoning~\cite{Feng2019,Fowl2021,Huang2021} use an iterative creation process, which is relatively inefficient.
Moreover, these studies use loss functions that require knowledge of the labels for specific images, which might not be desirable in privacy-sensitive scenarios.

In this work, to address these limitations, we propose ShortcutGen, a new generative poisoning method capable of generating sample-dependent poisoning shortcuts.
Once trained, the generator can generate perturbations for any given input image with only one forward pass, which is naturally more efficient than iterative methods.
In addition, the use of a generator eliminates the need for label knowledge during the creation process.
The main difference between our ShortcutGen and the only existing generative method, DeepConfuse, is that our ShortcutGen only trains the generator with a static discriminator, while DeepConfuse requires time-consuming alternating updates between the discriminator and generator.
This difference makes our ShortcutGen about $\times 100$ faster than DeepConfuse.

In particular, we demonstrate that using a well-trained discriminator, following recent advances on generative adversarial examples~\cite{Poursaeed2018,Naseer2019,Naseer2021},~\textit{cannot} work but a randomly-initialized discriminator is needed.
This insight is based on our new conjecture that a randomly-initialized discriminator provides a noisy mapping between images and labels, encouraging the generator to learn spurious shortcuts instead of generalizable features.
We also explore possible ways to further improve ShortcutGen.
Specifically, we show that using a simple augmentation strategy boosts the robustness of ShortcutGen against early stopping~\cite{Huang2021,Sandoval2022}, and that using composite perturbations from ShortcutGen with and without this augmentation yields better performance than existing methods in terms of the final validation accuracy, especially in the challenging, transfer scenario.
Finally, we provide additional insights into the working mechanism of ShortcutGen and its potential to generate poisons for unseen data based on learning a stronger generator with a more general latent representation space.
\begin{figure}[!t]
    \centering
    \includegraphics[width = \textwidth]{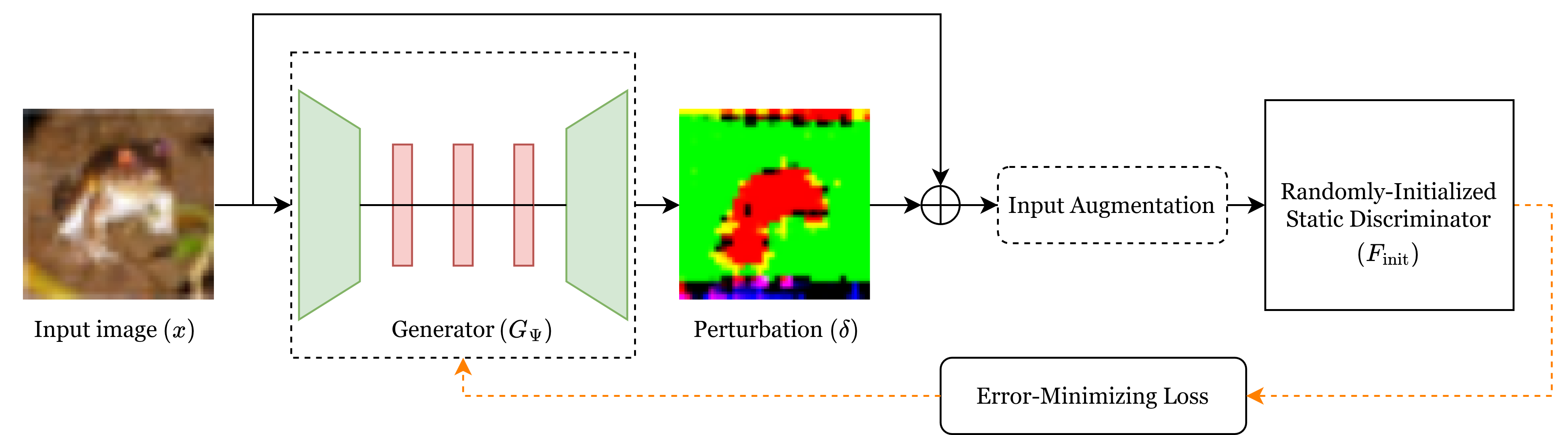}
    \caption{Training pipeline of our ShortcutGen for generating poisoning shortcuts.} 
    \label{fig: Training pipeline}
\end{figure}

\section{ShortcutGen}\label{sec: Method Section}
First, we briefly introduce the general formulation of the data poisoning problem studied in this work.
The general goal of data poisoning is to compromise the availability of a victim model by reducing its accuracy on clean test data\footnote{A thorough review of availability poisoning attacks can be found in \cite{Goldblum2022}.}. 
Following recent work~\cite{Feng2019,Fowl2021,Huang2021,Liu2021}, we achieve this goal by adding perturbations $\mathcal{P} = \{\boldsymbol{\delta}_i\}$ into the original (clean) training dataset $\mathcal{D}_o = \{(\boldsymbol{x}_i, y_i)\}$~\cite{Huang2021,Fowl2021} to obtain a poisoned training dataset $\mathcal{D}_p = \{(\boldsymbol{x}'_i, y_i)\}$. 
Specifically, the generated perturbations are sample-dependent, i.e., $\boldsymbol{x}'_i=\boldsymbol{x}_i+\boldsymbol{\delta}_i$, and bounded by $||\boldsymbol{\delta}_i|| \leq ~\epsilon=8/255$.


Next, we present technical details of ShortcutGen, which is inspired by existing work on generative adversarial perturbations \cite{Poursaeed2018,Naseer2021,Naseer2019}.
Fig. 1 presents the training pipeline of our ShortcutGen.
ShortcutGen learns a ResNet-based generator $\mathcal{G}_{\Psi}$, which takes as input a specific image sample $\boldsymbol{x}_i$ and outputs its corresponding bounded perturbations $\boldsymbol{\delta}_i$, against a discriminator $F_{\theta}$.
During training, only the generator is updated but the discriminator is static.
Specifically, ShortcutGen uses an error-minimizing loss~\cite{Huang2021} that aims to associate shortcut features with the true class of each image.
The generator is trained on the clean dataset $\mathcal{D}_o$ using the loss function Eq.~\ref{eq: Natural Extrapolation Architecture Loss}.
\vspace{-1cm}
\begin{multicols}{2}
  \begin{equation}
    \min_{\Psi} \mathcal{L} (F_{\theta}(\mathcal{G}_{\Psi}(\boldsymbol{x}) + \boldsymbol{x}), y)
    \label{eq: Natural Extrapolation Architecture Loss}
  \end{equation}
  \break
  \begin{equation}
    \min_{\Psi}\mathcal{L} (F_{\textrm{init}}(\textrm{Aug}(\mathcal{G}_{\Psi}(\boldsymbol{x}) + \boldsymbol{x})), y)
    \label{eq: POSTAUG objective}
  \end{equation}
\end{multicols}
We adopt the cross-entropy loss for $\mathcal{L}$ and use the Change of Variables~\cite{Carlini2017} to softly clip the perturbations for more accurate gradients.

\noindent\textbf{Random Discriminator.} We find that directly adopting the framework in~\cite{Naseer2019}, thus using a well-trained discriminator, can only reduce the model accuracy by 2\%.
We assume a well-trained discriminator provides the possibility of using highly-generalizable (yet not robust) features as the shortcuts.
To address this problem, we propose to instead use a randomly-initialized discriminator ($F_{\textrm{init}}$), which helps the generator learn spurious shortcuts, which are hardly generalizable. 
This is supported by the fact that random networks can detect random, foreground objects but have low discriminative capability \cite{Cao2022}.
Using a random discriminator can also simulate the early training stage of the victim model, thus it can generate faster learned poisons that should be more effective~\cite{Sandoval2022}.



\noindent\textbf{Input Augmentation Against Discriminator.}
Another important change we made beyond the framework in~\cite{Naseer2019} is to augment the input of the discriminator for improving the robustness of ShortcutGen against a potential early stopping defense~\cite{Huang2021,Sandoval2022}.
This is motivated by our finding that a victim model trained with data augmentations is capable of maintaining the model accuracy in the early training stage, in contrast to a victim model without data augmentations.
Our input augmentation is expected to reduce the impact of the victim model's data augmentations, relative to the literature, because it forces the generator to learn augmentation-invariant perturbations.

Overall, the loss function in Eq.~\ref{eq: Natural Extrapolation Architecture Loss} is modified to obtain Eq.~\ref{eq: POSTAUG objective} by integrating a random discriminator $F_{\textrm{init}}$ and input augmentation $\textrm{Aug}(\cdot)$.
We also combine the perturbations from the augmented and non-augmented variations (using maximal pixel values) to further boost the performance in terms of final validation accuracy, while remaining competitive in the peak validation accuracy.

\section{Results}
\label{sec: Results}
We first present the evidence of ShortcutGen's training efficiency. As can be seen from Table~\ref{tab:Times sample wise}, our ShortcutGen is roughly 100 times more efficient than DeepConfuse~\cite{Feng2019}, the only existing generative poisoning method.
\begin{table}[H]
  \centering
  \small  
 
    \begin{tabular}{l|c}
  
     Method  & Training Time\\
     \hline  
     DeepConfuse~\cite{Feng2019} & 10 days \\
     \hline
     ShortcutGen no Aug  & ~2 hours \\  
     ShortcutGen & ~2.5 hours \\
    \end{tabular}%
      
    \caption{Training time of compared generative poisoning frameworks for the training set of CIFAR10~\cite{CIFAR10} using the original code and hyperparameters. All methods were trained on the same hardware, which included 1 GeForce RTX 2080 Ti GPU.}
   \label{tab:Times sample wise}%
\end{table}%
Then, we compare the performance of our ShortcutGen (SG) to the only existing generative poisoning attack method, DeepConfuse (DC) \cite{Feng2019}, and also other state-of-the-art methods: Error-Minimizing (E-Min) \cite{Huang2021} and Error-Maximizing (E-Max, targeted) \cite{Fowl2021}.

\begin{figure}
\begin{floatrow}
\capbtabbox{%
          \resizebox{0.38\textwidth}{!}{
    \begin{tabular}{l|cc}

     Method  & Final & Peak (\# epoch) \\
    \hline
         Clean & 94.09 & 94.17 (58) \\

        \hline

     E-Min~\cite{Huang2021}  & 22.16 $\pm$ 0.33 & \textbf{24.3 $\pm$ 0.93} (28) \\
     
     E-Max~\cite{Fowl2021}     & 9.14 $\pm$ 0.44 & 50.52 $\pm$ 3.93 (2) \\

     DC~\cite{Feng2019}   & 13.34 $\pm$ 1.05 & 32.55 $\pm$ 3.34 (1) \\   
        \hline
     SG no Aug   & 10.45 $\pm$ 2.49 & 60.96 $\pm$ 2.97 (4) \\

     SG   & 36.04 $\pm$ 4.23 & 39.19 $\pm$ 3.67 (36) \\

     Combined  & \textbf{6.42 $\pm$ 0.42} & 41.02 $\pm$ 1.45 (2) \\

    \end{tabular}%
}
}{%
  \caption{Validation accuracy (\%) in terms of final (60 epochs) and peak (\# epoch) accuracy for a RN18 victim model on CIFAR10. Results are averaged over 5 runs.
  }\label{tab:main evaluation}%
}.
\capbtabbox{%
          \resizebox{0.55\textwidth}{!}{

    \begin{tabular}{l|ccccc}

     Method  & VGG11 & CNN & MN & WRN  \\
    \hline
        Clean & 91.42/91.42 & 67.56/67.67 & 77.37/77.61 & 95.64/95.64 \\
    \hline

     E-Min~\cite{Huang2021}     & 21.00/\textbf{23.06} & 21.61/\textbf{23.03} & 19.87/\textbf{28.02} & 20.56/23.48  \\
     
     E-Max~\cite{Fowl2021}       & 21.93/50.37 & 28.28/46.34 & 14.54/39.32 & 8.78/49.34  \\
 
     DC~\cite{Feng2019}     & 15.94/24.25 & 16.66/24.19 & 15.40/38.05 & 13.20/\textbf{18.94}  \\
        \hline
      SG no Aug & 20.26/52.64 & 20.31/46.10 & 17.90/55.91 & 13.61/53.01  \\

     SG      & 55.11/56.56   & 34.64/34.86 & 38.86/38.83 & 38.63/40.44 \\

       Combined      & \textbf{9.91}/30.83 & \textbf{13.58}/46.54 & \textbf{10.95}/43.18 & \textbf{5.87}/38.26  \\

    \end{tabular}%
    }


     
 



  %
}{%
  \caption{Final/peak transferability (\%) of a VGG11, MobileNetV2 (MN), or WideResNet (WRN) victim model when RN18 is adopted as the discriminator.} 
  
  \label{tab:Transferability}%
}
\end{floatrow}
\end{figure}
As can be seen from Table~\ref{tab:main evaluation}, SG (no Aug) reduces the final accuracy to random guessing (10 \% for CIFAR10) and further the best results are achieved by the combined method.
The superior performance in terms of final validation accuracy of the combined method also generalizes to the challenging, transfer scenario, as shown by Table~\ref{tab:Transferability}.
This performance gain may be explained by the fact that the combination inherits the strong shortcuts from no augmentation but with a lower start-point accuracy provided by augmentation.
In addition, as expected, incorporating the augmentation improves ShortcutGen on the peak accuracy.



\begin{figure}
\begin{floatrow}
\capbtabbox{%






        \centering
              \resizebox{0.4\textwidth}{!}{

    \begin{tabular}{l|cc}

      &    MNIST  & ImageNet \\
    \hline
     Clean         & 100.00/100.00 & 89.60/89.60  \\
    \hline
     DC~\cite{Feng2019}     & 97.39/98.66  & 84.40/86.20\\
         \hline

          SG no Aug     & 95.91/98.15 & \textbf{83.60}/\textbf{84.40}\\

          SG   &  99.03/99.06 & 84.80/85.80  \\

          Combined  & \textbf{86.59}/\textbf{97.55}  & 85.40/86.00 \\

    \end{tabular}%
    }
}{%
  \caption{Validation final/peak accuracy (\%) of a RN18 victim model trained on CIFAR10 poisons but tested on unseen (ImageNet or MNIST) data.}%
  \label{tab: Cross-Dataset}
}
\ffigbox{%
    \centering

\includegraphics[width = 0.5\textwidth]{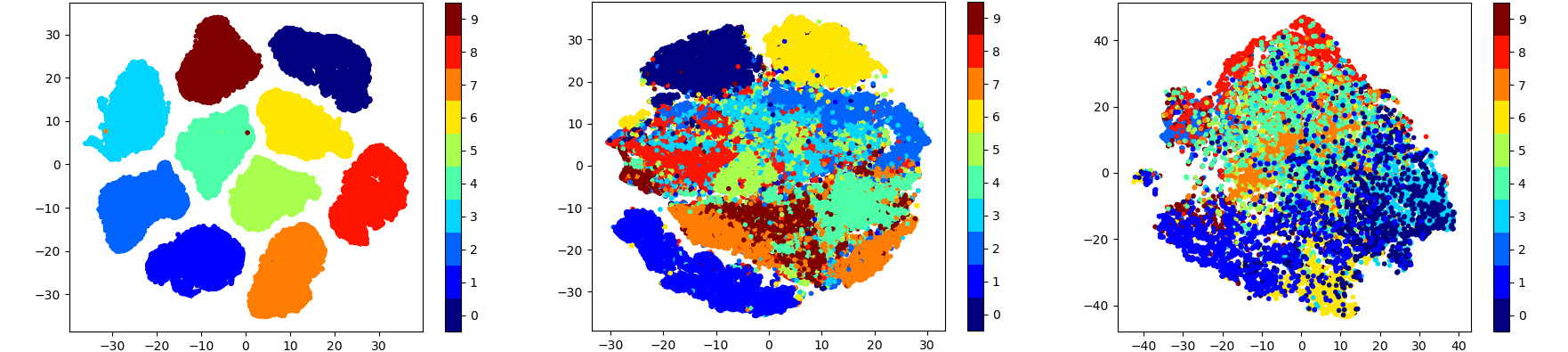}

}{%
  \caption{The t-SNE visualizations for the latent space of our generator (no Aug) on CIFAR10~\cite{CIFAR10} (left) , MNIST~\cite{MNIST} (middle) and subset of ImageNet~\cite{IMAGENET} (right).}%
  \label{fig: Random adversary workings shown}
}

\end{floatrow}
\end{figure}
In order to be truly useful in privacy-sensitive scenarios, generative poisoning should be successful when generating perturbations for data not seen during training.
As can be seen from Table~\ref{tab: Cross-Dataset}, our combined method has the potential for this purpose since it reduces the final model accuracy on ImageNet and MNIST.
Specifically, the results on MNIST are better than those on ImageNet.
This might be explained by the t-SNE~\cite{TSNE} visualizations of the last non-upsampling layer of the generator in Fig.~\ref{fig: Random adversary workings shown}, which shows that the feature space of the generator on MNIST is more separable than that on ImageNet, and fully-separable on CIFAR10.
These visualizations also highlight the working mechanism of our ShortcutGen and imply that learning a more general representation space allows it to work on unseen data.

\section{Conclusion}
\label{sec: Conclusion}
In this paper, we propose a novel generative poisoning method, ShortcutGen, for learning spurious shortcuts based on a randomly-initialized discriminator.
We demonstrate that integrating input augmentation makes ShortcutGen more robust against the early stopping defense, and combining an augmented and non-augmentation variant yields new state-of-the-art results in terms of the final validation accuracy, especially in the challenging, transfer scenario.
Finally, we uncover the working mechanism of ShortcutGen and demonstrate its potential to work on unseen data.


 
  
     
     
   
      

\bibliographystyle{splncs04}
\bibliography{ShortcutGen}

\end{document}